# Onion-Peeling Outlier Detection in 2-D data Sets

Archit Harsh
Master's Student
Electrical and Computer
Engineering Department
Mississippi State University
Starkville, MS, USA

John E. Ball
Assistant Professor
Electrical and Computer
Engineering Department
Mississippi State University
Starkville, MS, USA

Pan Wei
PhD Student
Electrical and Computer
Engineering Department
Mississippi State University
Starkville, MS, USA

## ABSTRACT

Outlier Detection is a critical and cardinal research task due its array of applications in variety of domains ranging from data mining, clustering, statistical analysis, fraud detection, network intrusion detection and diagnosis of diseases etc. Over the last few decades, distance-based outlier detection algorithms have gained significant reputation as a viable alternative to the more traditional statistical approaches due to their scalable, non-parametric and simple implementation. In this paper, we present a modified onion peeling (Convex hull) genetic algorithm to detect outliers in a Gaussian 2-D point data set. We present three different scenarios of outlier detection using a) Euclidean Distance Metric b) Standardized Euclidean Distance Metric and c) Mahalanobis Distance Metric. Finally, we analyze the performance and evaluate the results.

## 1. INTRODUCTION

Outlier detection is a critical step in a large number of data mining, data exploration, and data analysis tasks. Examples abound ranging from its use in medical diagnostics [1], image analysis [4], and network intrusion detection [10, 9] to its use as a pre-processing step for assessing the quality of data and as a precursor to various data mining algorithms that are heavily influenced by outliers.

Noted physicist Stephen Hawkins defined an outlier as "an observation which deviates so much from the other observations as to arouse suspicions that it was generated by a different mechanism." Similar outlier definition is proposed by NIST as "An observation that lies an abnormal distance from other values in a random sample from a population." [14]. Thus, a simple, mathematical definition of an outlier is an illusive concept. These definitions of outliers really leaves it up to the analyst (or a consensus process) to decide what will be considered outliers. To discriminate such outliers from normal observations, machine learning and data mining have defined numerous outlier detection methods, for example, traditional model-based approaches using statistical tests, or changes of variances and more recent distance-based approaches using $k$-nearest neighbors, clusters, or densities.

We focus in this paper on the distance-based approaches, which define outliers as objects located far away from the remaining objects. More specifically, given a metric space, each object $x \in M$ receives a real-valued outlier score $q(x)$ via a function $q : M \rightarrow \mathfrak{R}$; $q(x)$ depends on the distances between $x$ and the other objects in the dataset.

Then, the top-$k$ objects with maximum outlier scores are reported to be outliers. The outlier scores are determined based on the distance metric used for the analysis. Top-$k$ objects with the largest outlier scores are deemed as potential outliers.

In this paper, we present and implement a modified onion peeling algorithm to detect top-$k$ outliers in a Gaussian 2-D data set. The idea of onion peeling, or peeling in short, is to construct a convex hull around all the points in the dataset and then find the points that are on the edge of the convex hull. These points form the first 'peel' and are removed from the dataset. Repeating the same process gives more and more peels, each containing a number of points. We modified this basic idea to detect the $k$ largest outliers in a given 2-D Gaussian data-set. The choice of $k$ is influenced by the spatial geometry of the data-set and is user-defined. The convex hull is the smallest convex set that contains all of the points in the set.

The proposed algorithm works in two phases:

1. Find the convex hull using the onion peeling algorithm

2. Based on the convex hull and data-set, compute the number of outliers with the maximum distance from the centre of the data.

The major contributions of this paper are:

1. A detailed convex hull algorithm for finding potential outliers at the shallow layers is proposed. The proposed algorithm runs in linear time, which makes it efficient for computational purposes.

2. Application of the algorithm in Gaussian 2-D data-sets.

3. Comparison of performance using two widely known distance metrics: Euclidean and Mahalanobis Distance.

The paper is organized as follows: section II discusses the related background. Section III presents the modified onion peeling algorithm. Section IV presents the various distance metrics used in the algorithm. The simulation results are presented in Section V, and a theoretical analysis of the results is given in Section VI.

## 2. RELATED WORK

Distance-based techniques for outlier detection have gained significant reputation and have been the centre point of data-analysis, data-exploration and clustering tasks, due to their relatively non-parametric nature, scalability and simple implementation. In literature, there are three main definitions of outliers [13]:

1. Outliers are objects with fewer than $k$ neighbours in the database, where a neighbour is an object that is within a distance $R$ [6].

2. Outliers are the $n$ objects presenting the highest distance values to their respective $k^{th}$ nearest neighbour (the $k$-NN definition) [5].

3. Outliers are the $n$ objects presenting the highest average distance to their respective $k$ nearest neighbours [12].

All three definitions are in accordance with Hawking's definition, that is, the greater is the distance of the object to its neighbours, the more likely it is an outlier. The first definition originally proposed by Knorr and Ng [6] relies on both the definition of a neighbourhood $R$ as well as the number of neighbours $k$ in order to define an outlier. The flexibility of choosing $R$ is also an additional parameter which is user centric. Accordingly, over one hundred discordancy/outlier detection tests have been developed for different circumstances, depending on data distribution, distribution parameters (mean and variance), the number of expected outliers etc. However, all of those tests suffer from the following two problems. First, almost all of them are univariate, restricting them to perform in multidimensional datasets. Second, all of them are distribution-based, making them sometimes difficult to apply, especially when there is a small amount of data, and the distribution parameters would be difficult to assess accurately.

To overcome the aforementioned problems of distribution fitting and restriction to univariate data-sets, computational geometry inspired approaches for outlier detection have been developed [11]. In these approaches, based on the geometry of the data-set, data-objects are organized in layers in the data-space, with the expectation that shallow layers are more likely to contain outlying data objects than are the deep layers. Peeling is a well-known notion of depth studied in [7]. In this paper, we modified onion peeling algorithm to detect $k$ largest outliers at the edges of the data-set.

In the proposed work, 99% of the data-points constitute the normal set (no outliers) and the remaining 1% constitute the outlying set. Based on the onion-peeling definition, one can also establish the fact that the outlying set has the maximum volume enclosed by the hull as the potential outliers are found at the edges of the data-set.

# 3. ONION-PEELING ALGORITHM

The Onion-Peeling Algorithm behaves as follows:

Consider a set $S$ of $n$ points on a 2D plane. Compute the convex hull of $S$, and let $S'$ be the set of points remaining in the interior of the hull. Then compute the convex hull of $S'$ and recursively repeat this process until no more points remain. One ends up with a sequence of nested convex hulls, called the onion-peeling of S. The no. of points '$n$' in the set is denoted by |S|.

In this study, Graham's Scan algorithm [2] was utilized for computing the convex hull. Graham's Scan Algorithm is a widely used algorithm for computing convex hulls in linear time. It first explicitly sorts the points in O($nlogn$) and then applies a linear-time scanning algorithm to build the hull.

It works in three phases:

1. Find an extreme point. This point will be the *pivot*, is guaranteed to be on the hull and is chosen to be the point with the largest $y$ coordinate.

2. Sort the points in order of increasing angle about the pivot. The resulting polygon is usually a star-shaped polygon in which the pivot can see the whole polygon.

3. Build the hull, by marching around the polygon, adding edges when turning left, and back tracking when turning right.

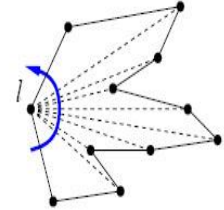

A simple polygon formed in the sorting phase of Graham's scan.

**Figure 1: Graham's scan sorting phase.**

Finally, to convert the polygon into convex hull, we apply the *three-penny algorithm* [Appendix A]. The scanning phase has the run-time complexity of O($n$) yielding an overall time complexity of O($n\ logn$). The detailed algorithm is provided in appendix A. This step formulates the first phase of our algorithm.

The first step works as follows:

1. Sort all points of S by x-coordinate.

2. Run Graham's Scan algorithm to compute the convex hull $H$ of $S$.

3. Remove the vertices of $H$ from $S$. If |S| >2, repeat step 2; else finished.

This step is crucial as it provides a basic understanding of the type of the dataset used for the analysis. By visual inspection of the hulls, one can estimate whether the underlying data-set contains outliers or not. This step also aids in estimating the no. of outliers in the data.

## 3.1 Onion Peeling Outlier Detection Algorithm

In this section, a modified onion peeling algorithm for detecting $k$ largest outliers randomly distributed in a 2-D space is presented.

The fundamental idea is that the largest outlier in the dataset will potentially be on the first peel based on Hawking's idea and onion peeling [11]. Hence, by inspecting the total distance of each point on the hull to all the other points, we can find the one with the largest total distance. Removing this point from the dataset and repeating this process gives new layers and new set of points. The algorithm is a recursive one depending on the number of outliers ($k$) and geometry of the data-set. The overall time complexity of this step is O($knd$) where $k$ is the number of outliers and $d$ is the chosen distance metric complexity.

We compared the performance of the algorithm using three different scenarios: a) Using default Euclidean distance metric b) using variance-standardized data before peeling and c) using the Mahalanobis distance metric.

### 3.1.1 Basic Definitions and Notations

**Definition 1:** Given a 2-D Gaussian Space, let '*n*' be the number of data-points randomly distributed in space. A potential outlier '*k*' is one which has the maximum distance from the mean of the data-set.

Definition 1 correlates with our underlying assumption and expectation that the outliers are more likely present in the shallow layers as compared to the depth layers. This definition is in accordance with Hawking's definition of outlier and the aforementioned outlier definitions.

The algorithm is a recursive algorithm. Based on the number of outliers '*k*' and '*n*', it scans the entire data-space using Graham's Scan algorithm and builds the convex hull. The algorithm calculates the outliers based on the distance metric, namely *Euclidean distance* and *Mahalanobis distance*. The objects with the maximum distance from the center are labelled as "potential outliers".

**Definition 2:** The output of the algorithm are outlier indices and volume. *OutlierIDXs* is defined as the indexes of rows in points that correspond to the outliers. *Volumes* is defined as the calculated volume of the convex hull.

Definition 2 explains the output parameters returned by the algorithm. The 'Volumes' parameter changes with each iteration as the algorithm converges.

This algorithm is fairly straight forward and easy to understand. One potential issue with this algorithm is its time complexity. However, we have optimized the algorithm implementation to reduce the complexity to O(*n logn*), where '*n*' is the number of items in the dataset, by applying Graham's Scan Algorithm [3]. The overall complexity of the algorithm is *O(nlogn) + O(knd)*. The entire algorithm runs in linear time which makes it efficient and suitable for intensive computational purposes.

The algorithm is implemented in MATLAB running on 2.4 GHz computer with 4GB of RAM.

The flowchart of the proposed algorithm is given as follows:

1. Provide the number of points randomly distributed in space.

2. Provide the number of estimated outliers, *k*.

3. Initialize the Graham's scan algorithm for computing the convex hull and volume computed in the process.

4. Based on the convex hull and distribution, run the outlier detection step by choosing the appropriate distance metric.

5. Calculate the distance of each point from the center.

6. Points that are furthest away from the data are labelled as outliers based on our "outlier definition". Convergence criterion is met when there are no set of points that can constitute a hull.

---

**Algorithm 1**: Onion Peeling Outlier Detection Algorithm

**Procedure:** Search for Outliers

**Inputs:** No. of points in the data *n*, Outliers to be identified *k*, Distance metric chosen: *EucDistance*, for Euclidean distance and *MahalaDistance* for Mahalanobis distance.

**Outputs:** *[OutliersIDxs, Volumes],* the outlier result set

**Define:** OutlierIDXs returns the indexes of rows in points that are potential outliers.

**Define:** Volumes contains the volume of the convex hulls measured in the process.

**Let:** Distance metric is a) Euclidean b) Mahalanobis.

**Begin:**

1: check the size of points.

  If *Size > k*

    Accept

  Else

    Error ("Size must be greater than outliers")

  end

2. Initialize the algorithm

  a) Run Graham's Scan Algorithm (given in appendix A)

    Get first hull

    Compute Volume

    Return [Volumes, hulls]

  b) Start removing points one at a time

    For all *k*

    Display ("Finding Outliers")

    *n* = size of the data)

3. Calculate distance with requested distance metric.

  Use equation (1) to (3)

4. Find the point with largest distance to all the points in the data

5. Store the point and the corresponding index in the memory.

6. Remove the current hull.

7. Compute new hulls.

8. Repeat steps 3-7 until convergence

9. Return the number of identified outliers and plot the result.

  end for

end

---

## 4. DISTANCE METRICS USED IN THE ALGORITHM

We will denote the distance between two objects $x$ and $y$ as $d(x, y)$, where $x$ and $y$ are $n$-dimensional vectors $x = (x_1, \ldots x_n)$ and $y = (y_1, \ldots y_n)$.

*A. Euclidean distance*

The Euclidean distance is given by

$$d(x, y) = \sqrt{(x_1 - y_1)^2 + \ldots (x_n - y_n)^2} \qquad (1)$$

This metric is the most widely used distance metric in most of the distance-based outlier detection algorithm and yield good results.

*B. Standardized Euclidean distance*

In some situations, values along the first dimension is relatively larger than in other dimensions, then the first dimension usually dominates the Euclidean distance. To avoid this unwanted situation, one solution is to weight each term in equation (1) with the inverse of the variance of that dimension.

$$d(x, y) = \sqrt{\frac{(x_1 - y_1)^2}{\sigma_1^2} + \ldots + \frac{(x_n - y_n)^2}{\sigma_n^2}} \qquad (2)$$

Where, $\sigma_i^2 (i = 1 \ldots n)$ is the sample variance in each dimension. This is often called the standardized Euclidean distance.

*C. Mahalanobis distance*

Mahalanobis distance takes variability into account that naturally occur naturally within the data. It is calculated from

$$d(x, y) = \sqrt{(x - y) \sum{}^{-1} (x - y)^T} \qquad (3)$$

Where, $\sum$ is the covariance matrix whose *(i, j)* entry is the covariance

$$\sum\nolimits_{i,j} = E[(X_i - \mu_i)(X_j - \mu_j)]$$

Where,

$$\mu_i = E(X_i)$$ is the expected value of the *i*[th] entry in X.

*D. Peeling with Euclidean Distance*

Peeling uses the Euclidean distance by default. The data-set is normalized before peeling to provide uniform weights in both *x* and *y* dimensions. The evaluation is performed on raw and normalized data-sets.

*E. Peeling with Mahalanobis Distance*

It is possible to use a completely different distance measure for calculating the total distance of a point on the hull to all the other points in the dataset. The motivation for using the Mahalanobis distance is rooted in the fact that it seeks to measure the correlation between variables and yield robust results.

In most 2-D scenarios, Euclidean distance neglects the weight of the *x* and *y* dimensions, since it treats each feature equally, thereby neglecting the overall weight of the variable. For instance, the outliers may tend to be more associated with the y dimension as compared to x or vice versa. In such scenarios,

Euclidean distance fails considerably as it relies on uniform distribution.

On the other hand, Mahalanobis distance considers the overall correlation of variables in the data and tend to identify outliers more naturally as compared to Euclidean distance.

## 5. SIMULATION RESULTS

The algorithm is implemented on a synthetic Gaussian 2-D data-set. The data-set is created with zero mean and unity variance across one dimension and zero mean and variance of 100 in second dimension. Outliers are plotted in the respective dataset using scatter plot showing the clustered data set, with a different colour and marker for each cluster. The desired outliers are fed to be 15 using three different scenarios to evaluate the performance of the algorithm.

The programming tool used to implement the algorithm is

MATLAB [8]. This is because MATLAB is a very powerful tool computing system for handling the calculations involved in scientific and engineering problems.

With MATLAB, computational and graphical tools to solve relatively complex science and engineering problems can be designed, developed and implemented.

*A) First scenario using Euclidean distance and raw data Set.*

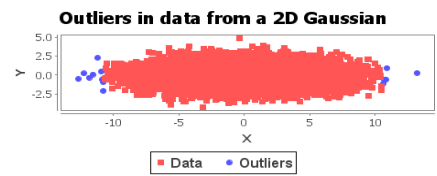

**Figure 2: outlier detection using raw data-set**

Figure 2 shows the results using Euclidean distance. As clearly evident, all the outliers are concentrated across one dimension which signifies the drawback of Euclidean distance metric in 2-D scenarios.

*B) Second scenario using standardized Euclidean distance*

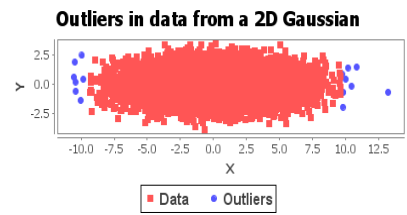

**Figure 3: outlier detection using standardized data-set**

Figure 3 shows the results using standardized Euclidean distance. The plots are almost identical to Figure 2. However, it reveals more outliers as compared to Figure 2. This might be due to the standardization/scaling of the data before peeling.

*C. Third scenario using Mahalanobis Distance*

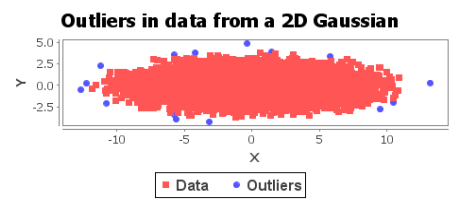

**Figure 4: outlier detection using Mahalanobis distance**

Figure 4 shows the results using Mahalanobis distance. As previously mentioned, the outliers are more naturally occurring in both dimensions. This is due to the suitability of this distance metric in 2-D scenarios.

The results in Figure 4 confirms our motivation of using Mahalanobis distance in comparison to Euclidean distance.

Simulations were performed 10 times for validating the performance of the algorithm. Table 1 demonstrates the number of common outliers in the top 6 runs for each different combination. The algorithm converged in a maximum of 10 iterations for each different run. As outlier detection approaches may account for errors, the performance threshold of the given algorithm is chosen to be 75%

**Table 1: Merit of the proposed algorithm**

| Accuracy | Threshold | Merit |
|---|---|---|
| (76-100)% | 75% | Good |
| 75% | 75% | Average |
| (1-74)% | 75% | Bad |

**Table 2: Common outliers in each method**

| Run | 1 | 2 | 3 | 4 | 5 | 6 |
|---|---|---|---|---|---|---|
| Peeling (Euclidean) | 9 | 8 | 0 | 9 | 7 | 9 |
| Peeling (Standardized) | 7 | 8 | 9 | 0 | 9 | 5 |
| Peeling (Mahalanobis) | 12 | 9 | 6 | 12 | 12 | 2 |

Table 1 validates the performance of the algorithm by considering three different cases. In case 1 and 2, a maximum of 9 outliers are commonly detected by the algorithm among the 15 outliers signifying an accuracy of 60%. In case 3, the algorithm detected 12 out of 15 outliers signifying an accuracy of 80% by using Mahalanobis distance metric.

Based on the performance threshold, the algorithm has a good performance in the third scenario where we used the Mahalanobis distance metric. This observation is critical in a sense that Mahalanobis distance takes the overall variation in the 2D data-set involving weights of both the dimensions.

## 6. THEORETICAL ANALYSIS:
Our major contribution in this work is the modification of onion peeling algorithm for detecting outliers since it is primarily designed for detecting edges in a convex hull.

In Table 1, we have validated the performance of the algorithm by considering three different cases. In case 1 and 2, a maximum of 9 outliers are commonly detected by the algorithm among the 15 outliers signifying an accuracy of 60%. In case 3, the algorithm detected 12 out of 15 outliers signifying an accuracy of 80% by using Mahalanobis distance metric. The reason for this improvement is possibly the choice of distance metric since Mahalanobis distance metric considers the overall variability in the data and gives precise results.

In figure 2 and 3, we can easily witness the similarity between the raw and standardized data set. Since, we considered Gaussian data-set, standardization doesn't make much difference. The prime observation is that outliers behave differently depending on the type of data-set. The evaluation in table 1 also demonstrates similar results.

In figure 4, we used Mahalanobis distance which accounts for the variability. Hence, the outliers are projected almost uniformly along both the dimensions which is an accurate representation of the outliers since in unsupervised learning, a user does not have any prior knowledge of the data-set and the nature of outliers.

We also observed that by changing the distance metric, the results seem to be more interesting in 2-D data-sets, as each dimension contribute to the potential outliers. Particularly, in this work, we considered Gaussian 2-D data set with high variance along the second dimension. Mahalanobis distance metric seems to be the potential candidate for detecting outliers since it accounts for the overall variability in the data-set.

## 7. CONCLUSION AND FUTURE SCOPE
In this paper, a modified onion peeling algorithm for the purpose of outlier detection in 2-D data-sets is presented. The performance of the algorithm is evaluated by considering three different scenarios. The algorithm works well for outlier detection and by changing the distance metric, we found that Mahalanobis distance metric suits well for 2-D data-sets in comparison to standard Euclidean distance due to its flexibility to account for variability resulting in 80% accuracy and 33.33% improvement in performance. We also observed that the nature of the outliers is highly correlated with the type of data-set used and the number of data-points.

In our future work, we will implement the algorithm on real data-sets. In general, Onion peeling is independent of the dimensions. So, it would be interesting to implement the algorithm on high- dimensional data-sets to validate its scalability.

thethird IEEE International Conference on Data Mining, page 601. Citeseer, 2003.

## 9. APPENDIX A:
## Graham's Scan Algorithm:

Graham's Scan Algorithm, first explicitly sorts the points in $O(nlogn)$ and then applies a linear-time scanning algorithm to finish building the hull.

The first step in this algorithm is to find the point with the lowest y-coordinate. We start the scan by finding the leftmost point l. Then, we sort the points in counterclockwise order around l. Any general purpose sorting algorithm can accomplish the task. We used *Heapsort* as our sorting algorithm. The time complexity of this step is $O(nlogn)$.

To compare two points p and q, we check whether the triple *l,p,q* is oriented clockwise or counterclockwise. Once the points are sorted, we connect them in counterclockwise order, starting and ending at l. The result is a simple polygon with n vertices.

To convert the polygon into a convex hull, we apply the following '*three penny algorithm*'.

We have three pennies, which will set on three consecutive vertices *p,q,r* of the polygon; initially, the pennies will be l and any two vertices succeeding l. We now apply the following two rules iteratively until a penny moves to l.

1. If *p,q,r* are in counterclockwise order, move the penny forward to the successor of r.

2. If *p,q,r* are in clockwise order, remove q from the polygon, add the edge pr, and move the middle penny backwards.

Whenever a penny moves forward, it moves onto a vertex that hasn't seen a penny before (except the last time), so the first rule is applied $n−2$ times. Whenever a penny moves backwards, a vertex is removed from the polygon, so the second rule is applied exactly $n − h$ times, where h is as usual the number of convex hull vertices. Since each counterclockwise test takes constant time, the scanning phase takes $O(n)$ time altogether.

The overall time complexity is as follows:

1. The sorting step takes $O(nlogn)$.

2. The scanning step takes $O(n)$.

3. The total time complexity is

$O(n)+O(nlogn)+O(n) = O(nlogn)$

The second phase of the outlier detection algorithm works by calculating the no. of outliers based on the chosen distance metric. The time complexity of this step is $O(knd)$ where k is the no. of outliers and d is the chosen distance metric.

Hence, the overall time complexity of the proposed approach is $O(nlogn)+O(knd)$.

The following figure demonstrates the three penny scanning step to build the convex hull.

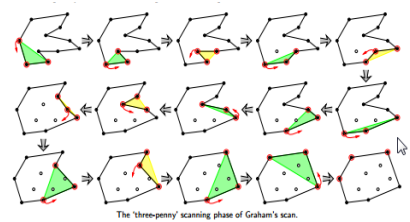

The 'three-penny' scanning phase of Graham's scan.

## PSEUDOCODE:

**Input:** A set of points **S** = {n = (n.x,n.y)}

    1. Select the rightmost lowest point $n_o$ in **S**
    2.Sort **S** radially (ccw) about $n_0$ as a center
{
    Use isLeft() comparisons
    For ties, discard the closer points
}
Let P[N] be the sorted array of points with P[0]=$n_0$

Push P[0] and P[1] onto a stack Ω

while i < N
{
    Let $P_{T1}$ = the top point on Ω
    If ($P_{T1}$ == P[0]) {
      Push P[i] onto Ω
      i++  // increment i
    }
    Let $P_{T2}$ = the second top point on Ω
    If (P[i] is strictly left of the line $P_{T2}$ to $P_{T1}$)
{
      Push P[i] onto Ω
      i++  // increment i
    }
    else
      Pop the top point $P_{T1}$ off the stack
}
**Output:** Ω = the convex hull of **S**.